# A COMPARATIVE STUDY OF SEVERAL ADPCM SCHEMES WITH LINEAR AND NONLINEAR PREDICTION[1]

*Oscar Oliva, Marcos Faúndez-Zanuy*
Escola Universitària Politècnica de Mataró
Universitat Politècnica de Catalunya (UPC)
Avda. Puig i Cadafalch 101-111, E-08303 Mataró (BARCELONA)
e-mail: faundez@eupmt.es http://www.eupmt.es/veu

## ABSTRACT

In this paper we compare several ADPCM schemes with nonlinear prediction based on neural nets with the classical ADPCM schemes based on several linear prediction schemes.
Main studied variations of the ADPCM scheme with adaptive quantization (2 to 5 bits) are:
-forward vs backward
-sample adaptive vs block adaptive

## 1. INTRODUCTION

Many speech coders are based on linear prediction coding. With LPC it is not possible to model the nonlinearities present in the speech signal, so there is a growing interest for nonlinear techniques.
This paper contributes to increase knowledge about the behaviour of the nonlinear prediction of speech based on a Multilayer perceptron neural nets.
Several works have been published revealing the better performance of nonlinear techniques but little attention has been dedicated to the implementation of the nonlinear model into real applications, so a lot of work must be done in this subject.
The summary of our previous work (available in pdf format in our web) is the following:
In [1] we studied a nonlinear prediction model based on neural nets. In [2] we proposed an ADPCM scheme with nonlinear prediction and a novel hybrid ADPCM-Backward scheme which combines linear and nonlinear prediction. In [3] we proposed an efficient algorithm for reducing the computational burden. In [4] we propose a sample adaptive scheme, and in [5] the hybrid prediction scheme is studied in detail.
We obtained that the ADPCM forward with NLPC scheme obtains similar performance than the LPC one but with one bit less in the quantizer. Thus, for instance the 40Kbps ADPCM-forward LPC is equivalent (similar SEGSNR) than the ADPCM-forward NLPC at 32Kbps. This implies that for the same bit rate, the NLP outperforms the LPC with 2.5 to 3 dB in SEGSNR (except for a quantizer of only 2 bits) for a wide range of frame length, but this study did not inlude the effect of weight quantization nor the additional information needed to send the weights of the MLP. This paper will deal the ADPCM forward scheme with quantized predictor coefficients.

The results of this paper have been obtained with the same database that our previous papers, but the sentences have been concatenated, rather than encoding each sentence alone.

## 2. ADPCM WITH LINEAR PREDICTION

The coefficients are updated based upon the previous transmitted information, so it is not necessary to send any information about the prediction coefficients.
Two different adaptation schemes have been studied:
a) sample adapative prediction coefficients via LMS (that is, the predictor coefficients are updated for each input sample)
b) block adaptive prediction coefficients via Levinson Durbin recursion (that is, the predictor coefficients are updated one time for each input frame).

Table 1 shows the results for the first scheme, and table 2 for the second one.

| Nq | 2 | 3 | 4 | 5 |
|---|---|---|---|---|
| Gp (dB) | 7.78 | 8.48 | 8.57 | 8.58 |
| SEGSNR (dB) | 13.9 | 19.51 | 24.39 | 29.14 |

*Table 1. ADPCMB-LMS-10*

| Nq | longframe | 8=1 | | 8=0.92 | |
|---|---|---|---|---|---|
| | | Gp | SEG SNR | Gp | SEG SNR |
| 2 | 50 | 7.37 | 14.09 | 7.72 | 14.27 |
| 3 | 50 | 9.13 | 19.71 | 9.38 | 19.9 |
| 4 | 50 | 9.76 | 24.07 | 10.08 | 24.42 |
| 5 | 50 | 10.21 | 28.54 | 10.5 | 28.99 |
| 2 | 100 | 7.94 | 14.55 | 7.93 | 14.41 |
| 3 | 100 | 9.58 | 20.15 | 9.46 | 20.04 |
| 4 | 100 | 10.27 | 24.77 | 10.23 | 24.97 |
| 5 | 100 | 10.72 | 29.38 | 10.71 | 29.62 |
| 2 | 200 | 7.61 | 14.28 | 7.38 | 14.03 |
| 3 | 200 | 9.03 | 19.95 | 8.84 | 19.71 |
| 4 | 200 | 9.67 | 24.74 | 9.59 | 24.69 |
| 5 | 200 | 10.08 | 29.45 | 10.03 | 29.47 |

*Table 2. ADPCMB-Levinson-Durbin-10*

Main results are:
- There is no significative improvement increasing the predictor order for the first scheme (sample adaptive).

[1] This work has been supported by the CICYT TIC97-1001-C02-02

- For the second scheme (block adaptive), a little improvement can be achieved using bandwith expansion. (The LPC coefficients are multiplied by $a'_k = \lambda^k a_k$, $k = 1,\ldots,P$).
- The block adaptive scheme outperforms the sample adaptive. See also section 6.2 of [6].
- For the sample adaptive scheme, the Gp remains approximately constant (except for Nq=2) while in the block adaptive scheme there is a significative increase for each additional residual signal quantization bit.

| Nq | longframe | $\delta=1$ | | $\delta=0.92$ | |
|---|---|---|---|---|---|
| | | Gp | SEG SNR | Gp | SEG SNR |
| 2 | 50 | 5.76 | 12.83 | 7.63 | 14.23 |
| 3 | 50 | 7.62 | 18.91 | 9.18 | 19.85 |
| 4 | 50 | 8.29 | 23.32 | 9.87 | 24.27 |
| 5 | 50 | 8.52 | 27.6 | 10.28 | 28.7 |
| 2 | 100 | 7.72 | 14.60 | 8.02 | 14.58 |
| 3 | 100 | 9.17 | 20.27 | 9.53 | 20.2 |
| 4 | 100 | 9.88 | 24.77 | 10.29 | 24.94 |
| 5 | 100 | 10.23 | 29.3 | 10.74 | 29.63 |
| 2 | 200 | 7.62 | 14.44 | 7.27 | 13.98 |
| 3 | 200 | 8.98 | 20.18 | 9.18 | 19.63 |
| 4 | 200 | 9.59 | 24.86 | 9.87 | 24.64 |
| 5 | 200 | 9.89 | 29.34 | 10.28 | 29.42 |

table 3: ADPCMB-Levinson-Durbin-25

## 3. ADPCM WITH NL PREDICTION

### 3.1 Sample adaptive scheme

In [4] we proposed a scheme for a sample adaptive nonlinear prediction, which is a generalization of the block adaptive system, with an increasing updating rate.

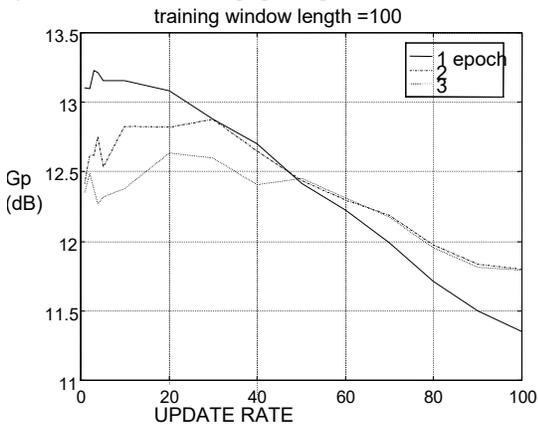

*Figure 1 Gp ADPCM-B Nonlinear prediction.*

Main results were:
a) The computational burden increases drastically.
b) The sample adaptive scheme (1 new predictor for each input sample) obtains a lower SEGSNR than the scheme with 1 new predictor for each 5 input samples, but this is due to the effect of the quantizer (compare figures 1 and 2. Altough Gp increases, SEGSNR is lower when updating more frequently the predictor coefficients). This is because the quantizer based on multipliers is not suitable for this application.

Although there are minor differences in linear prediction between forward (with unquantized coefficients) and backward schemes (about 0.5 dB in SEGSNR), in the nonlinear prediction schemes, this differences increase to 3 dB, so it is more interesting the study of the ADPCM forward (with quantized coefficients) than in the linear case.

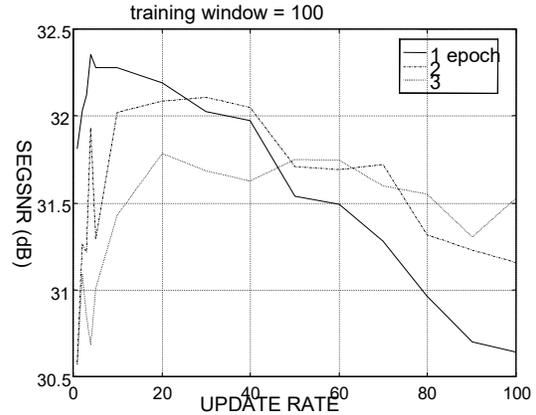

*Figure 2 SEGSNR ADPCM-B Nonlinear prediction.*

### 3.2 Forward scheme

This section is devoted to the study of the ADPCM forward scheme with quantized predictor coefficients. In order to obtain an optimal quantizer, it is important to know the statistical distribution of the predictor coefficients.

Figure 3 shows the weights from the input to the first hidden neuron

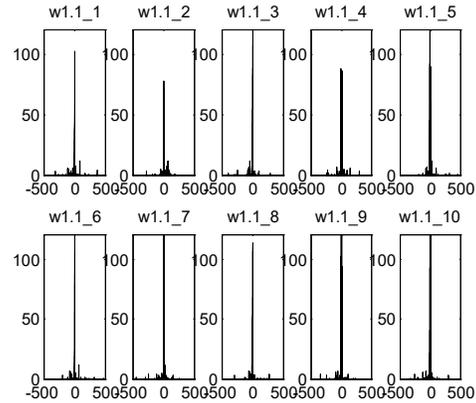

*Figure 3 Histogram of the weigths for the 1st hidden neuron*

Altough we do not include the histograms of the whole set of parameters (10x2+2 weights + 3 bias), from figures 3 and 4 is possible to deduce that if scalar quantization is chosen, several quantizers must be designed for each kind of parameter. That is, one quantizer for:
1. Hidden neuron weigths (w1.0_x, w1.1_x; x =1,...,10)
2. Hidden neuron bias (b1.0,b1.1)
3. Output neuron weights (w2_1,w2_2)
4. Output neuron bias (b2)

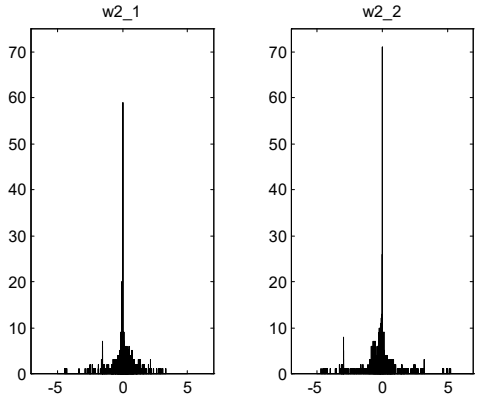

*Figure 4 Histogram of the weigths of the output neuron*

Obviously, other alternatives are possible, such as the use of a vector quantizer to exploit the redundancies between the different weights of the same predictor. In the linear case, this alternative yields better results. For instance, [7] references that a VQ at 10 bits/frame is comparable in performace a 24 bits/frame scalar quantizer. When the predictors are nonlinear, it is not trivial to obtain a vector quantizer. Altough several algorithms have been proposed [8],[9], when the dimensionality of the vectors is high, the computational burden and the number of training sentences is very high.

One of the goals of this paper is to study the robustness of the nonlinear predictive models against parameter quantization, so we think that the scalar quantization is enough interesting in order to study the behaviour of the NL predictor, rather than proposing an optimal ADPCM scheme.

A first experiment consists on the uniform scalar quantization using 10 bits/parameter (250 bit/frame). The parameter of design is the percentage of samples outside the bounds of the quantizer. Figure 5 shows the results as function of this parameter for a frame length of 200 samples and Nq=5 bits/sample (number of bits for prediction error quantization)

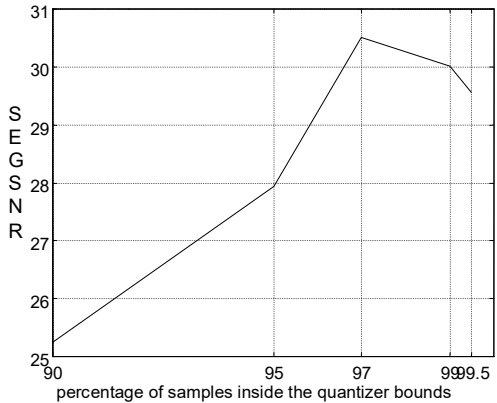

*Figure 5 Uniform scalar quantization of the NL predictor*

Obviously, better results are obtained using non-uniform scalar quantization. For this purpose, we have used the algorithm described in [10] that basically consists on adjusting the quantization levels such us the occupancy probability of all the steps of the quantizer is the same.

If the same number of bits is assigned to each parameter of the neural net, the results as function as the number of quantization bits for the residual signal is shown in figure 6. The lines are for 6 to 10 bits/parameter. We have limited the number of bits to 10 because in our previous work we found that the ADPCM forward with unquantized coefficients is comparable in performance to the backward ADPCM scheme using one bit less (for instance 48Kbit/s ADPCM forward similar to 60 Kbit/s ADPCM backward). Thus, no improvement will be obtained if the number of bits devoted to encode the predictor coefficients is greater, so in the following we will assume that the maximum number of bits for parameter is 10.

The next question is to study which parameters are more important (less robust to quantization). For this purpose, we have assigned several combinations of bits to the 4 groups of parameters described previously (w1,b1,w2,b2), so 7 10 7 10 means 20x7+2x10+2x7+1x10 = 184 bits/frame.

Figure 7 shows the SEGSNR for several combinations. It is easy to see that there are optimal combinations, and that the most important parameters are the related with the output.

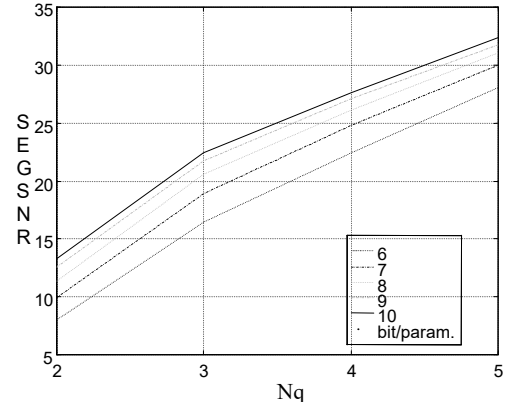

*Figure 6 SEGSNR in ADPCM-F. NL prediction*

Choosing the optimal conditions of figure 7, we can obtain several bit rates. In order to compare it with the bit rate of the ADPCM backward, the number of bits for each sample or prediction error must be added to the number of bits for encoding the parameters of the neural net. Figure 8 shows the overall bit rate against the SEGSNR in the forward ADPCM scheme, in addition to the SEGSNR of the backward scheme.

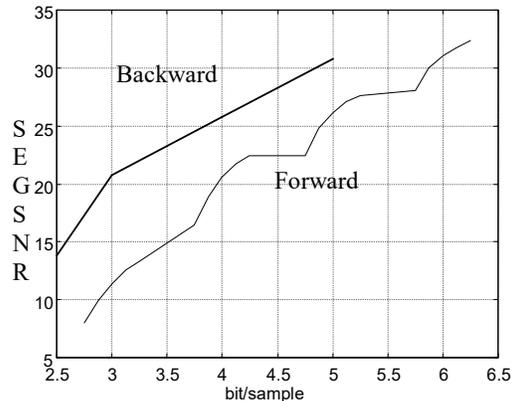

*Figure 8 SEGSNR vs overall bit rate.*

## 4. CONCLUSIONS

Main conclusions are:
- Altough there is a great improvement in SEGSNR over linear prediction in the forward scheme (the predictor coefficients are computed over the same frame to be encoded), this could only be achieved with a transparent quantization, that using our alternative requires a high number of bits. Really, our system requires 1 bit extra per sample in order to achieve the same SEGSNR than the backward scheme. Thus, it is not a valid alternative to the backward scheme proposed in our previous papers.

- Different number of bits are needed for quantizing the weights of the different layers of the neural net, because similar distributions exists for weights of the same layer and different neuron, in the forward scheme.
- In our experiments we have found that the neural net is less robust to quantization than the LPC coefficients, unless the quantization procedure is taken into account during the computation of the neural net. Really, it is preferable a mismatching between training and test conditions, than a disagreement between the computed weights and the used weights (due to a quantization process).
- In robustness with nonlinear prediction, one of the key factors is to achieve generalization capability (compromise between overtraining and adjustment of the neural net to the training problem), so perhaps the neural net must be undertrained, in a similar way than we found in our previous work about ADPCM with backward adaptation [2].

# REFERENCES

[1] M. Faúndez, E. Monte & F. Vallverdú "A comparative study between linear and nonlinear speech prediction". *Biological & artificial computation: from neuroscience to technology. IWANN-97.* pp.1154-1163. September 1997

[2] M. Faundez, Francesc Vallverdu & Enric Monte, "Nonlinear prediction with neural nets in ADPCM" ICASSP-98 ., Vol I, pp.345-348. SP11.3 Seattle, USA

[3] M. Faúndez, F. Vallverdú & E. Monte "Efficient nonlinear prediction in ADPCM". 5th International Conference on Electronics, circuits and systems, ICECS´98. Vol.3 pp.543-546. September 1998 Lisboa.

[4] M. Faúndez, O. Oliva "ADPCM with nonlinear prediction". EUSIPCO-98., Rodas pp 1205-1208.

[5] M. Faúndez , "Adaptive Hybrid Speech coding with a MLP/LPC structure". IWANN'99

[6] J. H. Chen "Low-delay coding of speech", in *Speech coding & synthesis*. Ed. Elsevier 1995, WB Kleijn & KK Paliwal edit.

[7] K.K. Paliwal, B. S. Atal "Efficient vector quantization of LPC parameters at 24 bits/frame". IEEE trans. on Speech & audio processing, Vol. 1 nº 1, January 1993.

[8] J. Thyssen, H. Nielsen, S. D. Hansen "Quantization of non-linear predictors in speech coding". ICASSP'95, pp.265-268.

[9] M. Faúndez "Speaker recognition by means of a combination of linear and nonlinear predictive models". EUROSPEECH'99, Budapest

[10] N. S. Jayant & P. Noll *Digital coding of waveforms: principles and applications* . Ed. Prentice Hall 1984

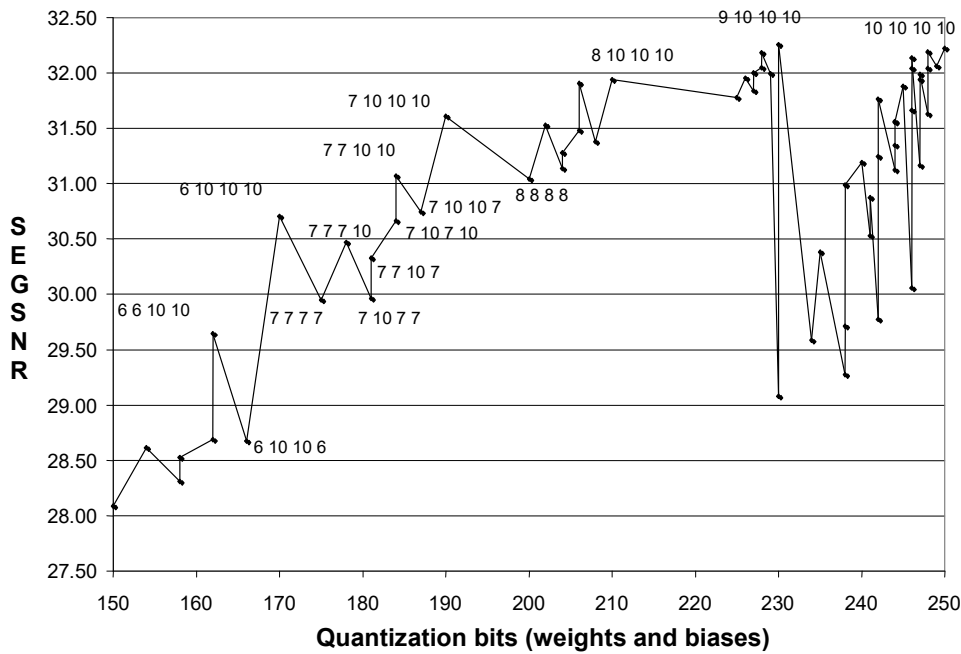

*Figure 7 Different bit assignments to each parameter of the NL predictor.*